\documentclass[11pt]{article}

\usepackage[final]{acl}

\usepackage{times}
\usepackage{latexsym}
\usepackage[T1]{fontenc}
\usepackage[utf8]{inputenc}
\usepackage{microtype}
\usepackage{inconsolata}
\usepackage{graphicx}
\usepackage{multirow}
\usepackage{spverbatim}
\usepackage{booktabs}
\usepackage{tabularx}
\usepackage{multicol}
\usepackage{enumitem}
\usepackage{float}
\usepackage{amsmath}
\usepackage{comment}

\title{UW-BioNLP at ChemoTimelines 2025: Thinking, Fine-Tuning, and Dictionary-Enhanced LLM Systems for Chemotherapy Timeline Extraction}

\author{
 \textbf{Tianmai M. Zhang\textsuperscript{*}},
 \textbf{Zhaoyi Sun\textsuperscript{*}},
 \textbf{Sihang Zeng\textsuperscript{*}},
 \textbf{Chenxi Li\textsuperscript{*}},
\\
 \textbf{Neil F. Abernethy},
 \textbf{Barbara D. Lam},
 \textbf{Fei Xia},
 \textbf{Meliha Yetisgen}
\\
 University of Washington
\\
 \small{
   \textbf{Correspondence:} melihay@uw.edu
 }
}

\begin{document}
\maketitle
\begin{abstract}
The ChemoTimelines shared task benchmarks methods for constructing timelines of systemic anticancer treatment from electronic health records of cancer patients. This paper describes our methods, results, and findings for subtask 2---generating patient chemotherapy timelines from raw clinical notes. We evaluated strategies involving chain-of-thought thinking, supervised fine-tuning, direct preference optimization, and dictionary-based lookup to improve timeline extraction. All of our approaches followed a two-step workflow, wherein an LLM first extracted chemotherapy events from individual clinical notes, and then an algorithm normalized and aggregated events into patient-level timelines. Each specific method differed in how the associated LLM was utilized and trained. Multiple approaches yielded competitive performances on the test set leaderboard, with fine-tuned Qwen3-14B achieving the best official score of 0.678. Our results and analyses could provide useful insights for future attempts on this task as well as the design of similar tasks.
\end{abstract}

\section{Introduction}

\def\thefootnote{*}\footnotetext{These authors contributed equally.}\def\thefootnote{\arabic{footnote}}

Electronic health records (EHRs) contain rich information on treatment courses, but extracting temporal relationships is challenging due to variability in care and linguistic complexity~\citep{Olex-2021, gholipour2023extracting}. Oncology regimens often deviate from planned schedules through dose changes or delays, with such modifications usually recorded only in unstructured notes that require chronological alignment~\citep{wang2020achievability}. Clinical narratives add further difficulty with relative or vague time expressions and inconsistent date formats~\citep{sun2013temporal,sun2015normalization}. Even experts may diverge in interpreting underspecified terms, making accurate normalization and sequencing a persistent challenge for clinical NLP systems.

The ChemoTimelines shared task\footnote{\url{https://sites.google.com/view/chemotimelines2025}\label{fn:task_website}}~\citep{yao-2024-overview, yao-2025-overview} was created to benchmark systems for constructing systemic anticancer treatment (SACT) timelines directly from EHR notes. It consists of two subtasks. In subtask 1, besides the raw EHRs, gold standard annotations of treatment events (EVENTs) and time expressions (TIMEX3s) for each patient EHR note are provided, and the task is to determine temporal relations between them on the patient level. In subtask 2, the task is to extract the patient-level treatment timeline with only the raw EHR notes available. We focus on subtask 2 to provide insights into an end-to-end treatment timeline extraction system.

Large language models (LLMs) demonstrate superior comprehension and information extraction ability, and were widely used in the previous year of the challenge~\citep{haddadan-etal-2024-lailab,zhang-etal-2024-nyulangone}. Without dedicated prompt engineering and chain-of-thought reasoning~\citep{wei2023chainofthoughtpromptingelicitsreasoning}, zero-shot prompting on LLMs has shown poor performance~\citep{zhang-etal-2024-nyulangone} in the timeline extraction task.  
Domain-adapted fine-tuning has proven effective for SACT timeline extraction, with models like Flan-T5-XXL \citep{chung2022scalinginstructionfinetunedlanguagemodels} and PubMedBERT \citep{Gu_2021} achieving strong results \citep{haddadan-etal-2024-lailab, tan-etal-2024-kclab}. However, these approaches have predominantly utilized older or smaller-scale architectures, such as BART~\citep{lewis2019bartdenoisingsequencetosequencepretraining} and Flan-T5-XXL~\citep{chung2022scalinginstructionfinetunedlanguagemodels}, and predicted timelines based on sentence-level contexts. Recent studies on scaling laws suggest that leveraging larger powerful models with rich context presents a clear opportunity for further improvement~\citep{kaplan2020scalinglawsneurallanguage}. In parallel, pipeline systems—which first extract events with a curated dictionary and then identify relations \citep{haddadan-etal-2024-lailab,wang-etal-2024-wonder}—have been developed but typically show inferior performance to end-to-end systems. Despite integrating external knowledge, the pipeline approach may still be suboptimal.

Building on previous efforts, we explore a variety of strategies to fill the gaps. 
First, to analyze the impact of LLM-based reasoning, we compare a baseline prompting system with a reasoning system. Second, to rethink the impact of external knowledge, we design a dictionary-enhanced extraction approach. Finally, to explore multiple training strategies, we conduct supervised fine-tuning (SFT) and direct preference optimization (DPO) on the latest LLMs. Our fine-tuned Qwen3-14B system wins first place in the challenge leaderboard. We provide several novel insights into the task that may inform future attempts on this task, as well as the design of similar tasks.

\section{Problem Formulation}
\label{sec:methods}
The SACT timeline extraction task for each patient is formulated as extracting $m$ triplets $\mathcal{T}=\{\text{<}s_j, r_j, t_j\text{>}\}_{j=1}^m$ from a series of $n$ clinical notes $X=\{x_1, ..., x_n\}$ of the patient, where $s$ indicates a SACT entity, $t$ is a TIMEX, and $r$ indicates the relation between $s$ and $t$ selected from BEGINS-ON, ENDS-ON, and CONTAINS-1. Following practices of last year's teams~\citep{haddadan-etal-2024-lailab}, we used note-level gold-standard relation annotations on the training set as the training data for our systems.

Detailed descriptions of the task framework and the dataset can be found on the shared task's website\footref{fn:task_website} or in the overview paper~\citep{yao-2025-overview}. In short, the dataset covers three cancer types (breast cancer, melanoma, and ovarian cancer) and was split by the task organizer into a training set (69 patients, 2,910 note files), a development set (27 patients, 1,272 note files), and a test set (53 patients, 2,121 note files). Teams participating in the shared task received the annotated training and development sets for the development of their systems. The unannotated test set was released a few days before the submission deadline for teams to run their systems and submit predictions.

To extract triplets $\mathcal{T}$ from clinical notes $X$, previous patient-level approaches~\citep{zhang-etal-2024-nyulangone} directly processed the entire $X$, which may overwhelm the LLM, while the sentence-level approach~\citep{haddadan-etal-2024-lailab} separately processed sentences in each $x_i$, which may lack global context. In contrast to these approaches, we leverage a note-level approach that splits the entire task into two steps: (1) \textbf{note-level extraction}: extracting triplets $\mathcal{T}_i$ from individual notes $x_i$, with or without format postprocessing, and (2) \textbf{timeline aggregation}: normalizing the TIMEXs and aggregating $\{\mathcal{T}_i\}_{i=1}^n$ into a patient-level timeline $\mathcal{T}$.
This setting allows decoupling of LLM extraction performance from the final timeline-level performance, enabling us to evaluate and optimize the methods for each step. 
LLM-based methods for step 1 are described in Section~\ref{sec:note_level_methods}, and the aggregation method for step 2 is explained in Section~\ref{sec:aggregation_method}.

\section{Note-Level Extraction}
\label{sec:note_level_methods}
We compared 5 different strategies for the note-level extraction task, providing insights from various aspects. We further included an ensemble method in our challenge submission to probe the relationship between note-level extraction and timeline aggregation.
\subsection{Prompting Baseline}
The baseline approach uses prompt-based, one-pass LLM inference. A prompt template (Appendix~\ref{sec:prompts}) was carefully curated based on the task definition and provided note-level gold annotations, encompassing detailed task instructions, in-context examples, and formatting requirements for a structured output. Each clinical note was appended to the prompt without preprocessing. LLMs generate extracted chemotherapy events $\mathcal{T}_i$ from each note $x_i$ as a JSON array. 

\subsection{Thinking}
Recent advances have shown improved reasoning and end-task performance when enabling a chain-of-thought (CoT) before generating answers~\citep{wei2023chainofthoughtpromptingelicitsreasoning}. In light of this, we enabled the thinking mode of the models in the prompting baseline using the same prompt to explore whether CoT could improve the timeline extraction. 

During error analysis, we observed text span discrepancies between LLM extractions and note-level gold-standard annotations, which sometimes resulted in false negatives in exact match evaluation. Therefore, we further designed the following postprocessing rules for our prediction submission based on the thinking method: (1) for SACT names containing "chemo", remove all descriptors before them, such as "adjuvant" and "neoadjuvant"; (2) for SACT names combined with a slash (e.g., "Doxorubicin/Cyclophosphamide"), split them into separate events; (3) remove unnecessary words in time expressions, such as "approximately", "about", "around", and "in". We do not include the postprocessing step in development set evaluation results (Table~\ref{tab:dev_results}) for a fairer initial evaluation.

\subsection{Dictionary-Enhanced Extraction}
We rethought and adapted the approach used by the LAILab team in last year’s Task 2~\cite{haddadan-etal-2024-lailab}, structuring it into a three-step pipeline.

\textbf{Step 1: Dictionary-based chemotherapy event extraction.} Given a clinical note, we first applied a self-constructed chemotherapy dictionary for keyword matching. All matches were tagged with \texttt{<e>} and \texttt{</e>}. The dictionary was built from three sources: (1) HemOnc.org\footnote{\url{https://hemonc.org/wiki/Main_Page}}
, where we created separate dictionaries for breast cancer, melanoma, and ovarian cancer including regimen names, drug names, and abbreviations; abbreviations with only two letters were removed to reduce false positives (e.g., "AT"); (2) generic mentions such as "chemotherapy" and "chemo" from the baseline system\footnote{\url{https://github.com/HealthNLPorg/chemoTimelinesEval}}; and (3) annotated chemotherapy mentions from the training and development sets of Subtask 1. No test set annotations were used. Only the drug names were incorporated into the dictionary; no labeled spans or relations were carried over to Subtask 2. The complete dictionary is provided in Appendix~\ref{sec:dictionary}. 

\textbf{Step 2: LLM-based double checking and augmentation.} Sentences containing dictionary tags were passed to the Qwen-3 Thinking model for verification, which reduced false positives and recovered false negatives. The prompt template is in Appendix~\ref{sec:prompts_llm_verification}.

\textbf{Step 3: Context-enhanced relation extraction.} For each verified sentence, we constructed a window of the anchor sentence plus its preceding and following sentences. This context was fed into Qwen-3 for generating chemotherapy–time relation triplets. The motivation for using a local window was efficiency: fewer than 6\% of sentences in the dataset contain SACT annotations (Table~\ref{tab:anno_stats}), and chemotherapy events and time expressions generally appear within two consecutive sentences. Based on these observations, we modified the baseline system’s prompt for local sentence-level relation extraction to improve efficiency (Appendix~\ref{sec:sentence-level_relation_extraction}).

\subsection{Supervised Fine-Tuning (SFT)}
Motivated by LAILab's success in using SFT in the previous challenge~\citep{haddadan-etal-2024-lailab}, we performed SFT to adapt LLMs for note-level extraction using the provided gold annotations. Input prompts were structured using the same template as our prompting baseline. For the training targets, the output for each note was serialized into a JSON object containing a list of dictionaries. Each dictionary represented a single extracted event with three mandatory keys: "SACT", "relation", and "time". 

Our SFT approach differs from the method proposed by LAILab~\citep{haddadan-etal-2024-lailab} in three key aspects. First, their method operates at the sentence level, providing the model with only a target sentence and its immediate neighbors as context. In contrast, our note-level approach allows the model to leverage the richer contextual information present in the entire clinical note. Second, for output generation, they employed a specialized triplet linearization algorithm~\citep{huguetcabotnavigli2021rebelrelation}. We adopt a potentially more flexible strategy by serializing the extracted relations into a structured JSON object. Finally, while their best performance was achieved by finetuning Flan-T5-XXL~\citep{chung2022scalinginstructionfinetunedlanguagemodels}, we scale up to a 14B-parameter model from Qwen3~\citep{qwen3-technical-report}, a more recent and advanced model family.

\subsection{Direct Preference Optimization (DPO)}
Recent work suggests that models trained via SFT tend to memorize the training data, while subsequent training with reinforcement learning can enhance generalization and alignment with human preference~\citep{chu2025sftmemorizesrlgeneralizes}. Motivated by this, we framed the note-level extraction task as a preference alignment problem. Specifically, we defined the preference as: (1) the extraction is expected to align with the style in gold annotations and (2) the note-level extraction may favor outputs with higher recall over precision, operating on the assumption that the downstream timeline aggregation process would manage deduplication and resolve conflicting extractions.

To implement this, we employed an iterative DPO approach to construct a preference dataset and refine the policy model~\citep{zhangonline,tu2025enhancingllmreasoningiterative,rafailov2024directpreferenceoptimizationlanguage}. First, we warmed up a policy model by training it for 5 epochs via SFT. Next, to generate preference pairs, we used this initial model to produce 8 candidate outputs for each instance in the training set. For each set of candidates, we identified the output with the highest recall as the chosen response ($y_w$) and the one with the lowest recall as the rejected response ($y_l$). This process yielded a preference dataset of pairs where the chosen and rejected responses were distinct. 
We then further trained the SFT warmup model using DPO~\citep{rafailov2024directpreferenceoptimizationlanguage} on this dataset.

\subsection{Ensemble Method}
For the ensemble method, we concatenated note-level predictions generated by three models: SFT, SFT + DPO, and Thinking + Postprocessing. These combined predictions were then passed through the normalization and aggregation pipeline to produce final patient-level timelines. We excluded the dictionary-enhanced extraction approach from the ensemble because its pipeline differs substantially from the other methods and introduces systematic false positives due to keyword matching.

\section{Timeline Aggregation}
\label{sec:aggregation_method}
Similar to how the task organizer constructed gold timelines automatically~\citep{yao-2024-overview}, all note-level system outputs underwent two subsequent steps, normalization and aggregation, to obtain patient-level timelines as final outputs. 

In the normalization step, time expressions in note-level outputs were converted into standardized ISO time using the CLUlab's \texttt{Timenorm} synchronous context-free grammar module \cite{bethard-2013-synchronous}. The original \texttt{Timenorm} was written in Scala, we reproduced its core functions in Java. The "DOCTIME" of each note was used as a temporal anchor for relative time expressions extracted from that note. Such "DOCTIME" was identified via a regular expression that detects eight consecutive digits in the note text. Relative time expressions that could not be normalized by \texttt{Timenorm} were discarded. 

Normalized events were then de-duplicated and aggregated using the official aggregation script (\texttt{docker\_output\_to\_timeline.py}) provided by the task organizer\footnote{\label{fn:evalcode}\url{https://github.com/HealthNLPorg/chemoTimelinesEval}}.

\section{Experimental Setup}
\subsection{Models}
Our experiments for note-level event extraction utilized the following open-source LLMs: Qwen3 series of general-purpose dense models \citep{qwen3-technical-report} from 4B to 32B, plus a mixture-of-experts model Qwen3-30B-A3B (2507), and Google's specialized model for medicine, MedGemma-27B \citep{medgemma-technical-report}. All models were obtained from Hugging Face.

We used the \texttt{vllm} package under Python 3.10 for LLM inference, and \texttt{LlamaFactory} for LLM fine-tuning. Sampling parameters for LLM inference followed the setting recommended by the Qwen3 team: \texttt{temperature=0.6}, \texttt{top\_p=0.95}, \texttt{top\_k=20}, and \texttt{min\_p=0}. The default maximum output length (\texttt{max\_tokens}) was set to 4,096. Up to 4 NVIDIA A100 GPUs were utilized for either model inference or model training.

For dense Qwen3 models, the thinking mode was enabled by setting the \texttt{enable\_thinking} parameter to True, and \texttt{max\_tokens} was changed to 20,480 to allow complete outputs; for Qwen3-30B-A3B (2507), the non-thinking model and the thinking model are two separate models.

\begin{figure}[ht]
    \centering
    \includegraphics[width=\linewidth]{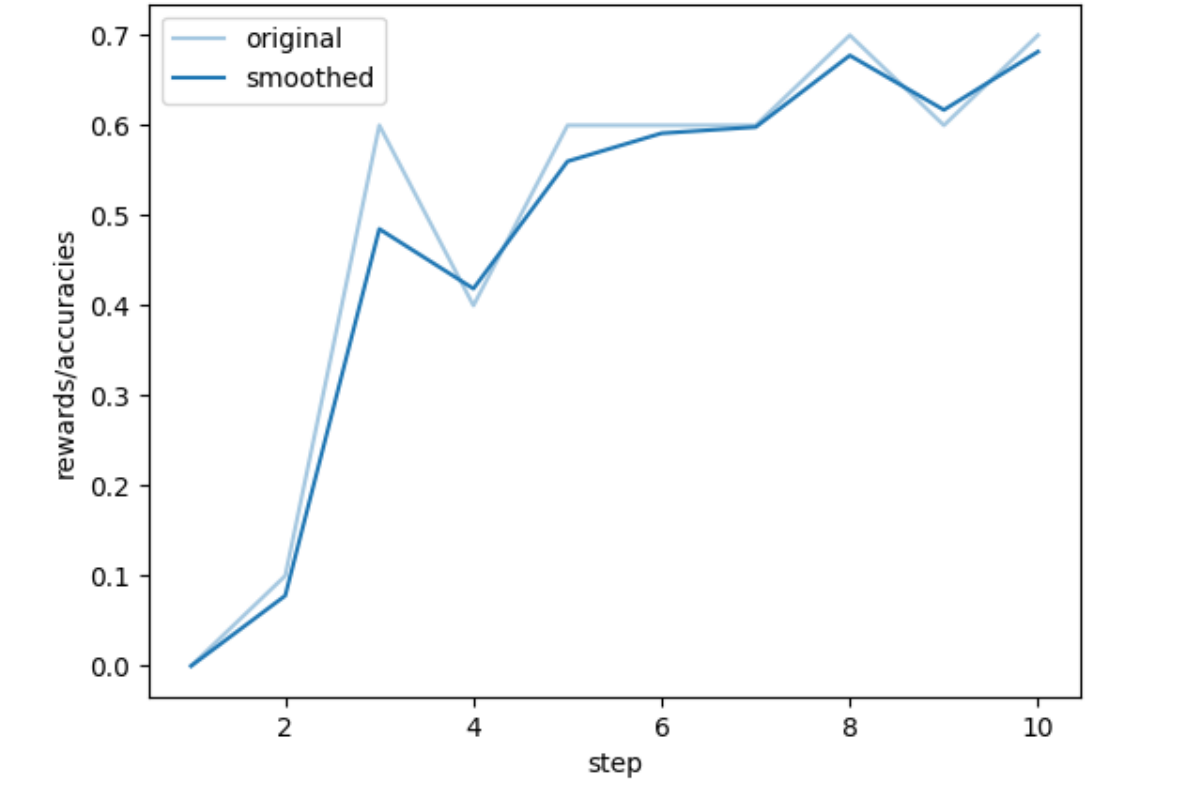}
    \caption{DPO reward accuracy curve of Qwen3-14B.}
    \label{fig:dpo}
\end{figure}

For SFT, we turned off the thinking mode of the Qwen3 model and employed LoRA for parameter-efficient finetuning of the model over 10 epochs. For DPO, we obtained preference datasets of 9 pairs for Qwen3-14B, 27 pairs for Qwen3-8B, and 30 pairs for Qwen3-4B. We trained with DPO for 10 epochs. Despite the small sample size, we observed a consistent improvement in reward accuracy during training (Figure \ref{fig:dpo}), which aligns with recent studies demonstrating effective reinforcement learning from a limited number of samples~\citep{wang2025reinforcementlearningreasoninglarge}.
\subsection{System Evaluation}
\label{sec:evaluation}
As instructed, system performance was assessed using the strict matching criterion, where all components of a predicted triplet must exactly match the corresponding gold standard triplet to be counted as correct. For the development set, evaluation was performed locally using the official evaluation script (\texttt{eval\_timeline.py}) provided by the shared task organizers\footref{fn:evalcode}. Since all pipelines followed 2 major steps as described in Section~\ref{sec:methods}, we additionally calculated note-level micro precision, recall, and F1 as intermediate metrics for LLM extraction performance. Evaluation on the test set was performed by the task organizer and announced using the leaderboard on the shared task website\footref{fn:task_website}.

Two types of patient-level F1 scores were calculated by the official evaluation script: Type A, which includes all patients regardless of whether they have gold-standard timelines, and Type B, which includes only patients with confirmed chemotherapy timelines \citep{yao-2025-overview}. The official score is the average of the Type A and Type B F1 scores, where each patient’s score is computed individually and then averaged across patients.

\section{Results}

\subsection{Development Performance}

Table~\ref{tab:dev_results} shows both note-level and timeline-level evaluation results of each method-model combination. Major findings are as follows.

First, better note-level performance is generally associated with, but does not necessarily indicate, better final performance on the timeline level, which highlights \textbf{the crucial role of normalization and aggregation}. For example, under the thinking approach, Qwen3-30B-A3B achieved much better micro precision, recall, and F1 than Qwen3-14B and Qwen3-32B, but their official scores were almost the same. This could be attributed to the deduplication of repeated events from different notes during event aggregation, and time expressions that \texttt{Timenorm} was unable to handle. Another noteworthy case is Qwen3-14B's exceptionally high official score under the baseline approach. Similar to other models, its note-level extractions contain a substantial proportion of irregular time expressions (see Section~\ref{sec:error_analysis} for examples), but those events were luckily discarded by \texttt{Timenorm}, leading to a high F1 score.

Second, \textbf{14B might be the best dense model size for this task}. Under both the baseline and thinking approaches, the 14B model outperformed other dense models on both the note level and the patient level. In light of this, we applied dictionary-based and fine-tuning-based methods to models up to 14B. As expected, Qwen3-14B consistently outperformed its 8B and 4B siblings. 

Third, \textbf{fine-tuning a dense model reliably yielded the largest performance gain, while a thinking mixture-of-experts model performed comparably}. After introducing rule-based postprocessing (described in Section~\ref{sec:note_level_methods}), Qwen3-30B-A3B's official F1 score on the development set improved from 0.596 to 0.625 (Table~\ref{tab:test_results}). We attribute most of the performance gain of thinking to the self-checking behavior exhibited by the CoT, which improved both the precision and recall of note-level event extraction. For details and examples, please see Section~\ref{sec:error_analysis}.

Fourth, \textbf{the dictionary-enhanced method achieved the highest recall among all systems, and incorporating LLM verification further improved precision by filtering out false positives.} We examined the intermediate results of LLM verification on top of dictionary-based sentence tagging. As shown in Table~\ref{tab:dict_llm_dev} in Appendix~\ref{sec:supplementary_fig_tab}, dictionary tagging alone achieved nearly perfect recall across cancer types, but precision was lower. Adding LLM verification consistently increased precision for these sentences (e.g., breast: 0.732$\rightarrow$0.824; melanoma: 0.802$\rightarrow$0.830) while keeping recall near~1.0 (a small trade-off appears for ovarian, 1.000$\rightarrow$0.994, with F1 unchanged). Overall, the dictionary-based pipeline attains the best recall among all systems, and the second-best development performance, trailing only SFT.

\begin{table*}
\centering
\begin{tabular}{l|l|ccc|ccc}
\hline
\textbf{Method} & \textbf{Model} & \multicolumn{3}{c}{\textbf{Note-Level Micro}} & \multicolumn{3}{|c}{\textbf{Timeline-Level Macro F1}}\\
& & \textbf{Precis.} & \textbf{Recall} & \textbf{F1} & \textbf{Type A} & \textbf{Type B} & \textbf{Official}\\
\hline
Prompting & Qwen3-4B & .039 & .278 & .069 & .173 & .082 & .127\\
Baseline & Qwen3-8B & .040 & .283 & .070 & .060 & .103 & .082\\
& Qwen3-14B & .139 & .276 & \textbf{.185} & .466 & .370 & \textbf{.418}\\
& Qwen3-32B & .103 & .209 & .138 & .253 & .220 & .236\\
& Qwen3-30B-A3B & .068 & .243 & .106 & .104 & .178 & .141\\
& MedGemma-27B & .085 & .439 & .142 & .158 & .199 & .178\\
\hline
Thinking & Qwen3-4B & .338 & .382 & .358 & .471 & .378 & .424\\
& Qwen3-8B & .355 & .335 & .345 & .614 & .410 & .512\\
& Qwen3-14B & .517 & .346 & .415 & .676 & .515 & .595\\
& Qwen3-32B & .355 & .325 & .339 & .623 & .568 & \textbf{.596}\\
& Qwen3-30B-A3B & .600 & .468 & \textbf{.526} & .676 & .516 & \textbf{.596}\\
\hline
Dictionary + & Qwen3-8B & .294 & .509 & .372 & .689 & .468 & .578\\
Sentence-level & Qwen3-14B & .434 & .657 & \textbf{.522} & .729 & .536 & \textbf{.632}\\
\hline
SFT & Qwen3-4B & .379 & .507 & .434 & .651 & .473 & .562\\
& Qwen3-8B & .419 & .569 & \textbf{.483} & .650 & .542 & .596\\
& Qwen3-14B & .397 & .615 & \textbf{.483} & .711 & .577 & \textbf{.644}\\
\hline
DPO + SFT & Qwen3-4B & .390 & .483 & .431 & .670 & .435 & .553\\
& Qwen3-8B & .409 & .574 & .478 & .651 & .545 & .598\\
& Qwen3-14B & .401 & .620 & \textbf{.487} & .695 & .549 & \textbf{.622}\\
\hline
\end{tabular}
\caption{Development set performance. The best official score under each method is highlighted.}
\label{tab:dev_results}
\end{table*}

\subsection{Test Performance}
On the test set, our SFT approach (submission 1) attained the highest overall average score of 0.678. The SFT + DPO model (submission 2) closely followed with an average of 0.666. The thinking approach with postprocessing (submission 4) also performed competitively, reaching an average score of 0.644. The ensemble method (submission 5), which combined SFT, SFT + DPO, and thinking achieved an overall score of 0.603, which was lower than any of the individual model. This suggests that differences in error patterns limited the benefit of ensembling. The dictionary-enhanced sentence-level extraction (submission 3) produced weaker results, with an overall score of only 0.545, suggesting potential limitations in the term coverage of our SACT dictionary with respect to what appears in the test set. Together, test results again indicate that fine-tuning-based methods were the most effective in our experiments, while LLM thinking was also a competitive approach.

\begin{table*}[ht]
\centering
\begin{tabular}{l|l|c|c} 
\hline
\textbf{Submission \#} & \textbf{Method} & \textbf{Dev Official} & \textbf{Test Official}\\
\hline
Submission 1 & SFT & \textbf{.644} & \textbf{.678}\\
Submission 2 & SFT + DPO & .622 & .666\\
Submission 3 & Dictionary + Sentence-level & .632 & .545\\
Submission 4 & Thinking + Postprocessing & .625 & .644\\
Submission 5 & Ensemble of 1, 2, and 4 & .562 & .603\\
\hline
\end{tabular}
\caption{Development and test set performance of final submissions. The best scores are highlighted.}
\label{tab:test_results}
\end{table*}

\section{Error Analysis}
\label{sec:error_analysis}
We empirically investigated noteworthy errors made by our systems on the development set, aiming to inform both future system development and potential refinements to the challenge design in subsequent rounds. Gold-standard timelines of the test set are held private by the organizer to enable future versions of the shared task, hence we are unable to perform error analysis on the test set.

\subsection{Errors in Prompt-Based Extraction}
Under the prompting baseline, LLMs often extract medications and procedures that are not part of SACT (e.g., "Neupogen", "ProHance", "MRI") and irregular time expressions (e.g., "04/26/2012 at 12:13 PM", "4 cycles", "midway through chemo"), even when explicitly instructed not to. They also produce errors related to text span boundaries and formatting. This explains the overly low precision and recall values on the note level.

When thinking mode is enabled, we observed that Qwen3 models would spontaneously check whether each candidate event belonged to SACT (e.g., "\textit{Neulasta is a G-CSF, not an SACT, so it's excluded}") and whether its associated time expressions satisfied the extraction instructions (e.g., "\textit{'status post' refers to something that happened in the past but doesn't give an exact time}"), which significantly reduced false positives. This double-checking behavior also helped decrease false negatives (e.g., "\textit{Double-checking to make sure I didn't miss any hidden events. Maybe ...}").

Most remaining errors produced by the thinking models were commonly encountered by other methods, and are discussed in the following sections. A noteworthy category is incorrect inference caused by ambiguous language or formatting in clinical notes. For example, consider a note containing a table of medications administered on a given date with the SACT note "TRASTUZUMAB" followed by "None Entered"; based on the column names in the context, this means that the start or end date of the therapy is not entered. However, Qwen3-30B-A3B interpreted this to mean that even though the date was present, it should be excluded because the time was missing. We also observed confusion about whether to include scheduled events. Our manual inspection of the gold standard timelines revealed that scheduled events were inconsistently annotated. In addition, a large proportion of clinical notes in the dataset contained no gold annotation, and any LLM-extracted events from these notes would become false positives.

\subsection{Errors in Dictionary-Based Extraction}
The dictionary-enhanced extraction approach provided strong coverage of chemotherapy mentions but also revealed several important limitations. First, the main limitation came from false positives, which lowered overall precision. Because the method matched any token found in the dictionary, it might incorrectly identify unrelated terms as chemotherapy mentions. For example, the system recognized the word "FEC" in "Normal FEV1 and FEV1/FEC ratio" as the regimen consisting of Fluorouracil, Epirubicin, and Cyclophosphamide, although it was in fact a typographical error for Forced Vital Capacity (FVC).

Second, the system also suffered from false negatives when encountering typographical errors or abbreviations not present in the dictionary. For instance, the test set included terms such as "bev" for Bevacizumab and "interfuron" for interferon. These variants were not captured, leading to missed extractions. This limitation helps explain the performance gap between the development set, where dictionary coverage was stronger, and the test set, where more novel variants appeared. 

Finally, we observed interesting cases of internal inconsistency between the model’s intermediate reasoning and its final output. For instance, in the thinking process, the model may explicitly state that a tag such as \texttt{<e>tc</e>} should be removed, but in the final output the tag still appears (see Appendix~\ref{sec:example_of_inconsistency} for an example). This mismatch suggests that controlling the alignment between reasoning and output remains a challenge for dictionary-enhanced extraction with LLMs. It also points the way to potentially useful future work in explainable AI to use reasoning traces to better understand how LLMs understand complex clinical notes.

\subsection{Errors in Training-Based Methods} 
As we pooled all available annotations into the same training set, the performance of our training-based model is sensitive to imbalances in the training data. For instance, the model often defaults to the generic CONTAINS-1 relation, misclassifying more specific BEGINS-ON and ENDS-ON relations. This tendency reflects a class imbalance where CONTAINS-1 instances are overrepresented in the training set. Furthermore, we observed a notable performance degradation on melanoma notes compared to breast and ovarian cancer, which manifests as low precision. This may be a consequence of data skew, as our training set contains significantly fewer melanoma notes, potentially leading the model to overfit to the majority of cancer types.

\subsection{Errors in Normalization and Aggregation}
The normalization process relies heavily on heuristic rules in the \texttt{Timenorm} pipeline, which can both improve alignment with the gold standard and introduce systematic errors. In general, using \texttt{Timenorm} facilitates consistent normalization of relative time expressions, but we found instances where the output diverged from expected interpretations. For example, the expression "last week" relative to a document time of 2013-01-15 is normalized to "2013-01-08" in full date format (YYYY-MM-DD), whereas "next week" relative to 2013-07-23 is normalized to "2013-w31" in week format (YYYY-w\#\#). Likewise, expressions that specify only month and day can be incorrectly anchored to the previous year. If the document time is 2013-02-10, the expression "January 9" is normalized as "2012-01-09" rather than the correct "2013-01-09". Such inconsistencies suggest that while Timenorm is powerful, it may require task-specific adjustments to handle edge cases in clinical timelines.

In aggregation, the lack of entity consolidation introduces redundancy and inconsistency across patient timelines. For example, the same chemotherapy drug can appear under slightly different surface forms, such as "il2", "il-2", and "interleukin-2", all linked to the same date and relation. Similarly, regimen-level mentions can coexist with individual drug mentions. The gold standard may annotate both "AC-T" as a regimen and its components Adriamycin (A), Cyclophosphamide (C), and Taxol (T), leading to multiple overlapping entries. 

A further source of discrepancy arises from how start and end events are aligned within the same timeline. When both BEGINS-ON and ENDS-ON relations are identified for the same drug on the same date, our system retains both events for completeness, whereas the gold timelines may arbitrarily keep only one. For example, in the gold timeline, Cabotaxol and Taxol are annotated as \texttt{[cabotaxol, BEGINS-ON, 2012-01-12]} and \texttt{[taxol, ENDS-ON, 2011-12-15]}. The complete representation, however, should include both start and end events for each drug, i.e., 4 events in total.

\section{Conclusions and Discussion}
\label{sec:discussion}
Extracting clinical events from unstructured notes has always been a challenging task~\citep{Olex-2021}. Under the ChemoTimelines 2025 shared task framework, our work explores several approaches based on modern and emerging model training and inference techniques. Major findings are as follows: \begin{enumerate}[noitemsep, topsep=5pt, leftmargin=*]
\item The aggregation of note-level events into patient-level timelines is crucial for the final performance of a system.
\item Fine-tuning a dense model, especially of size 14B, reliably yielded the largest performance gain, while a thinking mixture-of-experts model performed comparably.
\item The dictionary-enhanced method achieved the best recall, while LLM verification improved precision by reducing false positives.
\end{enumerate}

We found that the dictionary-based approach offered both efficiency and interpretability, while still maintaining acceptable performance despite some information loss at the sentence level. Instead of reasoning over all notes in the development set, the method reduced the burden by restricting LLM verification to a much smaller number of candidate sentences flagged by the dictionary, plus context-enhanced sentences for relation extraction. This substantially lowered input token volume, reasoning time, and computational cost. Although focusing on sentences inevitably sacrifices some contextual information compared with note-level extraction, the resulting performance remained strong, supported by very high recall from dictionary tagging and improved precision from LLM verification. Moreover, the transparent matching rules enhance interpretability and facilitate systematic refinements, such as synonym expansion or ontology-based extensions. Together, these features make the dictionary-based pipeline a lightweight, resource-efficient, and interpretable complement to learning-based systems.

The comparable performance of the training-free LLM thinking approach and fine-tuning-based methods suggests a potential cost-effectiveness trade-off for this specific task. Once fine-tuning is supported by a sufficient amount of high-quality data, it is capable of yielding a trustworthy performance gain while maintaining the speed of direct output generation. In contrast, CoT thinking, as a core component of recent inference-time scaling techniques for LLMs, is characterized by its higher latency at test time, despite that it requires much less data annotation in the development phase. Given the substantial performance gain from LLM thinking in our experiments (and potential benefits to explainability), we recommend that future attempts on similar tasks consider including it as a baseline method, especially in consideration of the high cost of EHR annotation by human experts.

The ensemble method did not lead to performance gains. Instead, the overall score was lower than any of the individual models. This suggests that errors from different systems tend to accumulate when combined, and these mistakes cannot be effectively corrected through the normalization and aggregation pipelines. As a result, simple ensembling is not a viable strategy for this task.

We conceived several other methods that were not implemented due to time and resource constraints, and we hope providing them here may benefit clinical timeline extraction. First, our current prompt-based approaches (both non-thinking and thinking) utilized static ad hoc in-context examples. Including dynamically-retrieved training examples related to test time queries has the potential to further improve performance. 
Second, current training methods do not have an explicit reasoning process before generating the extractions. Future methods may synthesize reasoning data through rejection sampling~\citep{yuan2023scalingrelationshiplearningmathematical} or apply reinforcement learning~\citep{deepseekai2025deepseekr1incentivizingreasoningcapability} for better performance.

The chemotherapy events in the training data represented the specific EHR documentation style of the source facilities and systems. The more general task of extracting clinical event timelines may involve a diversity of local documentation styles, event sources (e.g., treatment, laboratory, billing, etc.), and levels of standardization.

Additional insights into the capabilities and weaknesses of various LLM-based strategies might be obtained with introspection into performance against specific evaluation data examples, additional layers of case review with expert oncologists, and testing with the newest generation of emerging LLMs.

\section*{Limitations}
Our methods were highly customized to the ChemoTimelines challenge, hence our findings may not generalize well to other clinical extraction tasks. Due to time, resource, and privacy constraints, we did not assess a full range of contemporary open- and closed-source LLMs (e.g., larger Qwen3 models, the Llama series, GPT series, etc.), therefore our findings may not generalize. MedGemma was also the only medicine-specialized LLM included in our experiments. Although a general-purpose LLM combined with a tailored aggregation pipeline was sufficient for this task, future work may benefit from models more familiar with clinical notes. For technical methods that we conceptualized but did not have a chance to implement and test, please refer to Section~\ref{sec:discussion}.

\bibliography{custom}

\appendix

\section{Prompt Templates}
\subsection{Prompt for Baseline and Thinking}
\label{sec:prompts}
You are an experienced medical annotator tasked with extracting systemic anticancer therapy (SACT) events from a given clinical note.\\
\\
WHAT A SACT IS:\\
SACT encompasses medications used in traditional cytotoxic chemotherapy, endocrine therapy, targeted therapy, and immunotherapy. SACT may appear in generic names (e.g., Anastrozole), brand names (e.g., Arimidex), or combined names (e.g., TCH). Non-specific SACT mentions such as "chemotherapy" or "chemo" should also be included. Exclude therapies, medications, and diagnostic procedures used not for anticancer purposes, such as dietary supplements and biopsies. Exclude therapy candidates that you don’t know what they are or aren’t sure if they are SACT.\\
\\
WHAT YOU SHOULD EXTRACT:\\
You should only extract SACT events that are explicitly associated with specific time expressions. Here are some format examples of time expressions you are expected to extract: "December 27, 2011", "May 21st, 2013", "7/20/2012", "today", "3 weeks ago", "1 year". Ignore nonspecific time mentions such as cycle or dose numbers (e.g., "cycle 1 of 6" and "1/6 dose") and ambiguous relative time (e.g., "midway through" or "at the same time as" another event). Exclude SACT mentions without an associated time.\\
\\
Additionally, for each event, select a relation label from BEGINS-ON, ENDS-ON, and CONTAINS-1 to indicate the relation between the SACT and its time based on the note’s language. CONTAINS-1 means the SACT happened at a specific time; if the note explicitly mentions the start or end of an event, use BEGINS-ON or ENDS-ON. For example, if the note says "She received Herceptin on May 21st, 2013", your extracted event will be "Herceptin", "CONTAINS-1", "May 21st, 2013"; if the note says "Start ipilimumab on today’s date", your extracted event will be "ipilimumab", "BEGINS-ON", "today".\\
\\
If a SACT is associated with multiple time points, extract them as separate events. For example, if the note says "Herceptin was initiated on 12/27/2011 and completed on April 10, 2012", you should extract two events: one is "Herceptin", "BEGINS-ON", "12/27/2011", and the other is "Herceptin", "ENDS-ON", "April 10, 2012". Similarly, if multiple SACTs are associated with the same date, you should also extract them as separate events.\\
\\
HOW YOU SHOULD FORMAT YOUR RESPONSE:\\
SACT names and their associated time expressions should be kept exactly as they appear, even if there is a typo. Do not alter them, normalize the time expression, or infer the exact date. For example, if a SACT event appears in the note as "Alibercept received yesterdat", your extracted event will be "Alibercept", "CONTAINS-1", "yesterdat".\\
\\
Do not combine SACT mentions that refer to the same therapy but appear in different names, even if one appears in parentheses as the alternative name for another; extract them separately. For example, if there are three SACT events, "il-2", "il2", and "interleukin-2", and all have corresponding time expressions, treat them as three separate SACT events; if the note says "TRASTUZUMAB (HERCEPTIN) received today", you should extract two events: one is "TRASTUZUMAB", "CONTAINS-1", "today", and the other is "HERCEPTIN", "CONTAINS-1", "today".\\
\\
If multiple SACTs are administered together, treat them as separate events. For example, for "Doxorubicin/Cyclophosphamide" you should extract 2 events, one for Doxorubicin and the other for Cyclophosphamide. However, if a SACT name is already a combined treatment name (e.g., TCH), treat it as a single event.\\
\\
Ignore supplementary descriptors of SACT names, such as dose (e.g., "high dose") and administration method (e.g., "IV"). For non-specific SACT like "chemotherapy" or "chemo", ignore their descriptors, such as "adjuvant" and "neoadjuvant".\\
\\
Your response must be a JSON array under the following schema:\begin{spverbatim}
{
  "type": "array",
  "description": "An array of SCAT events extracted from the clinical note.",
  "items": {
    "type": "object",
    "properties": {
      "SACT": {
        "description": "A SACT name extracted as it is.",
        "type": "string"
      },
      "relation": {
        "description": "The relation between the SACT and its associated time expression. Must be one of BEGINS-ON, ENDS-ON, and CONTAINS-1.",
        "type": "string"
      }
      "time": {
        "description": "The time expression associated with the SACT, extracted as it is.",
        "type": "string"
      }
    },
    "required": ["SACT", "relation", "time"],
  }
}
\end{spverbatim}
If there is no SACT event in the clinical note, return an empty array.\\
\\
Now, extract SACT events from the following clinical note:\\
\{note\}

\subsection{Prompt for LLM-based Chemotherapy Tag Verification}
\label{sec:prompts_llm_verification}
You are an experienced medical annotator tasked with verifying and extracting systemic anticancer therapy (SACT) mentions from a given clinical note. Some SACT candidates have already been tagged using a dictionary-based method.

SACT encompasses medications used in traditional cytotoxic chemotherapy, endocrine therapy, targeted therapy, and immunotherapy. An SACT mention may appear as a generic name (e.g., Anastrozole), a brand name (e.g., Arimidex), or a combined name (e.g., TCH). Non-specific chemotherapy-related SACT mentions like "chemotherapy", "chemo", "chemotherapy's", etc. And even mentions with typos like "chemotheray" should also be retained.

Your task is twofold:

1. Review the pre-tagged mentions ONE BY ONE and remove any incorrect tags caused by dictionary false positives.

2. Identify and tag any additional SACT mentions that are missing due to typos or uncommon abbreviations not found in the dictionary.

Extract each SACT mention exactly as it appears in the note, even if there is a typo; do not alter or normalize it. For example, if an SACT Aflibercept appears in the note as "Alibercept", your extracted SACT should be "Alibercept". Do not combine SACT mentions that refer to the same therapy but appear in different forms; extract them as separate mentions. For example, if there are three mentions: "il-2", "il2", and "interleukin-2", extract them all separately.

Ignore supplementary information such as dose, administration method, or diagnostic/therapeutic context not related to anticancer treatment. Exclude therapies, medications, or procedures used for non-cancer purposes, such as dietary supplements or biopsies.

You should remove or add tags in the raw text. Do not output any other text.

Both the input and the output should be put in "{ }". Please strictly follow the format of the output. You MUST only wrap the correct SACT mentions with <e> and </e> tags in your outputs. Do not add any other tags or quote marks.

For example, given the input text:

"{This is a sentence with both <e>correct SACT</e> and <e>wrong SACT</e> mentioned.}"

The expected output is:

"{This is a sentence with both <e>correct SACT</e> and wrong SACT mentioned.}"

Sometimes the input can be extremely long, like:

"{======Here are some background details about the patient========

This is a sentence with both <e>correct SACT1</e> and <e>wrong SACT</e> mentioned, and another <e>correct SACT2</e> mentioned.}"

The expected output is:

"{======Here are some background details about the patient========

This is a sentence with both <e>correct SACT1</e> and wrong SACT mentioned, and another <e>correct SACT2</e> mentioned.}"

Now, extract SACT events from the following sentences in a clinical note:
\subsection{Prompt for Context-Enhanced Sentence-Level Relation Extraction}
\label{sec:sentence-level_relation_extraction}
You are an experienced medical annotator tasked with extracting systemic anticancer therapy (SACT) events from a given clinical note.

WHAT A SACT IS:
SACT encompasses medications used in traditional cytotoxic chemotherapy, endocrine therapy, targeted therapy, and immunotherapy. SACT may appear in generic names (e.g., Anastrozole), brand names (e.g., Arimidex), or combined names (e.g., TCH). Non-specific SACT mentions such as "chemotherapy" or "chemo" should also be included. I have extracted all the SACT events for you between the tags <e> and </e> in my input, so you don't need to extract SACT yourself. 

WHAT YOU SHOULD EXTRACT:
You should do this step by step. First, identify all the SACT events between <e> and </e> in my input and ONLY focus on these SACT events. Then, exclude the SACT events if they are macro information rather than patient-specific information, or if they are negations of SACTs. Next, for each valid SACT event, extract specific time expressions that are explicitly associated with that SACT event. Here are some format examples of time expressions you are expected to extract: "December 27, 2011", "May 21st, 2013", "7/20/2012", "today", "3 weeks ago", "1 year". Ignore nonspecific time mentions such as cycle or dose numbers (e.g., "cycle 1 of 6" and "1/6 dose") and ambiguous relative time (e.g., "midway through" or "at the same time as" another event). Exclude time expressions that are not associated with the current SACT event. If there is not a time expression related to the current SACT event, then skip it and check the next SACT event. 

Additionally, for each event, select a relation label from BEGINS-ON, ENDS-ON, and CONTAINS-1 to indicate the relation between the SACT and its time based on the note's language. CONTAINS-1 means the SACT happened at a specific time; if the note explicitly mentions the start or end of an event, use BEGINS-ON or ENDS-ON. For example, if the note says "She received Herceptin on May 21st, 2013", your extracted event will be "Herceptin", "CONTAINS-1", "May 21st, 2013"; if the note says "Start ipilimumab on today's date", your extracted event will be "ipilimumab", "BEGINS-ON", "today".

If a SACT is associated with multiple time points, extract them as separate events. For example, if the note says "Herceptin was initiated on 12/27/2011 and completed on April 10, 2012", you should extract two events: one is "Herceptin", "BEGINS-ON", "12/27/2011", and the other is "Herceptin", "ENDS-ON", "April 10, 2012". Similarly, if multiple SACTs are associated with the same date, you should also extract them as separate events.

HOW YOU SHOULD FORMAT YOUR RESPONSE:
SACT names and their associated time expressions should be kept exactly as they appear, even if there is a typo. Do not alter them, normalize the time expression, or infer the exact date. For example, if a SACT event appears in the note as "Alibercept received yesterdat", your extracted event will be "Alibercept", "CONTAINS-1", "yesterdat".

Your response MUST be in a json format under the following schema:

 hn {["SACT event1", "relation1", "time expression1"], ["SACT event2", "relation2", "time expression2"], … }
 
\section{Chemotherapy Events Dictionary}
\label{sec:dictionary}
\subsection{Breast Cancer}

\raggedcolumns
\footnotesize
5-fu \\
a-cmf \\
a.c \\
a/c \\
abemaciclib \\
abraxane \\
ac \\
ac-cmf \\
ac-d \\
ac-h \\
ac-t \\
ac-th \\
ac-thl \\
ac-thp \\
ach \\
act \\
adriamycin \\
afc \\
afinitor \\
airuika \\
alimta \\
alpelisib \\
anastrozole \\
anthracycline \\
arimedex \\
arimidex \\
aromasin \\
aromatase inhibitor \\
at-cmf \\
atc \\
atezolizumab \\
avastin \\
bev \\
bevacizumab \\
bilateral oophorectomy \\
caf \\
camrelizumab \\
capecitabine \\
capivasertib \\
carbo \\
carboplatin \\
cbd \\
cef \\
cef-t \\
chemo \\
chemo therapy \\
chemo-rt \\
chemoembolization \\
chemorad \\
chemort \\
chemotherap \\
chemotherapeutic \\
chemotherapeutic \\
chemotherapies \\
chemotherapy \\
chemotherapy's \\
chemotheray \\
chidamide \\
cisplatin \\
cmf \\
cmf-e \\
cmf-h \\
cmft \\
cnp \\
cp-ac \\
cp-ddac \\
cp-ec \\
cvb \\
cyclophosphamide \\
cytoxan \\
d-ac \\
d-ac+bev \\
d-ec \\
d-fec \\
d-fec+bev \\
datopotamab deruxtecan \\
datroway \\
dcb \\
dda-ddt-ddc \\
ddac \\
ddac-ddt \\
ddac-ddth \\
ddac-pacph \\
ddac-t \\
ddac-th \\
ddac-thp \\
ddat \\
dde \\
dde-iddcmf \\
ddec-ddcmf \\
ddec-ddd \\
ddec-ddt \\
ddec-t \\
ddec-th \\
ddec-thp \\
ddfec-d \\
ddp \\
ddt \\
ddt-ddec \\
ddt-ec \\
ddth \\
docetaxel \\
docetaxol \\
doxil \\
doxorubicin \\
e-cmf \\
e-d \\
e-x \\
ec-cmf \\
ec-d \\
ec-ddt \\
ec-dt \\
ec-h \\
ec-p \\
ec-t \\
ec-th \\
ec-thp \\
ecd-gc \\
ech \\
ech-th \\
edc \\
ehp \\
elacestrant \\
ellence \\
endocrine therapy \\
enhertu \\
enzalutamide \\
ep-ddcmf \\
epidaza \\
epirubicin \\
eribulin \\
everolimus \\
exemestane \\
fac \\
fac-t \\
fac-th \\
fac-thp \\
fareston \\
faslodex \\
fec \\
fec-d \\
fec-h \\
fec-p \\
fec-t \\
fec-th \\
fec-thp \\
femara \\
fluorouracil \\
fulvestrant \\
gcb \\
gdoc \\
gemcitabine \\
gemzar \\
ghp \\
gnrh analogs \\
goserelin \\
h+d \\
halaven \\
herceptin \\
herceptin hylecta \\
ibrance \\
idd-etc \\
iddenpc \\
iddepc \\
inavolisib \\
irene \\
itovebi \\
ixabepilone \\
ixempra \\
javlor \\
kadcyla \\
keytruda \\
kisqali \\
l+t \\
lapatinib \\
letrozole \\
leuprolide \\
loqtorzi \\
lupron \\
lynparza \\
margenza \\
margetuximab \\
methotrexate \\
millipred \\
mitomycin \\
mitoxantrone \\
mmm \\
mtx \\
mutamycin \\
myocet \\
navelbine \\
neratinib \\
nerlynx \\
nolvadex \\
novantrone \\
np-ddac \\
np-ddec \\
np-ec \\
npc-ddec \\
npld \\
ofs \\
olaparib \\
orserdu \\
ovarian irradiation \\
paclitaxel \\
palbociclib \\
paraplatin \\
pcb \\
pembrolizumab \\
pemetrexed \\
perjeta \\
pertuzumab \\
phesgo \\
piqray \\
platinol \\
platinum \\
pld \\
prednisolone \\
pyrotinib \\
q2wk \\
ribociclib \\
s-1 \\
sacituzumab govitecan \\
t-ac \\
t-cef \\
t-ddac \\
t-ddec \\
t-dm1 \\
t-ec \\
t-fac \\
t-fec \\
t-h \\
t-t \\
tac \\
talazoparib \\
talzenna \\
tamoxifen \\
taxane \\
taxol \\
taxotere \\
taxtotere \\
tc \\
tc-h \\
tcbh \\
tch \\
tchp \\
tcyh \\
tecentriq \\
th-ac \\
th-ddac \\
th-ech \\
th-fec \\
thl \\
thp \\
toremifene \\
toripalimab \\
tpc \\
trastuzumab \\
trastuzumab deruxtecan \\
trastuzumab emtansine \\
trelstar la \\
triptorelin \\
trodelvy \\
truqap \\
tucatinib \\
tukysa \\
tx-cex \\
tykerb \\
v-fec \\
verzenio \\
vh-fec \\
vhp \\
vinflunine \\
vinorelbine \\
xeloda \\
xhp \\
xtandi \\
zoladex 

\subsection{Melanoma}
\raggedcolumns
\footnotesize
abc \\
abraxane \\
afiblercept \\
aflibercept \\
alfa-2b interferon \\
alflibercept \\
alibercept \\
alpha 2b interferon \\
alpha interferon \\
alpha-2b interferon \\
alpha-2b interferon \\
alpha-2binterferon \\
atezolizumab \\
avastin \\
bevacizumab \\
binimetinib \\
braftovi \\
carboplatin \\
chemo \\
chemo therapy \\
chemo-rt \\
chemorad \\
chemoradiation \\
chemort \\
chemotherap \\
chemotherapeutic \\
chemotherapeutic \\
chemotherapies \\
chemotherapy \\
chemotherapy's \\
chemotheray \\
cisplatin \\
cnp \\
cobimetinib \\
complete resection \\
contego \\
cotellic \\
cpb \\
cvd \\
dabrafenib \\
dacarbazine \\
docetaxel \\
dtic \\
eldisine \\
encorafenib \\
fotemustine \\
gleevec \\
hepzato kit \\
il 2 \\
il-2 \\
il2 \\
imatinib \\
imlygic \\
inteferon \\
interferon \\
interferon \\
interleukin \\
interleukin 2 \\
interleukin-2 \\
ipilimumab \\
keytruda \\
kimmtrak \\
kolupin \\
koselugo \\
leukine \\
lifileucel \\
loqtorzi \\
mekinist \\
mektovi \\
melphalan \\
methotrexate \\
muphoran \\
nivolumab \\
opdivo \\
opdualag \\
paclitaxel \\
paraplatin \\
pembrolizumab \\
platinol \\
proleukin \\
sargramostim \\
selumetinib \\
tace \\
tafinlar \\
talimogene laherparepvec \\
tasisulam \\
taxol \\
taxotere \\
tebentafusp \\
tecentriq \\
temodar \\
temozolomide \\
tils \\
toripalimab \\
trametinib \\
tunlametinib \\
vaccinia \\
vaccinia virus \\
vaccinia virus \\
vemurafenib \\
vindesine \\
yervoy \\
zelboraf 

\subsection{Ovarian Cancer}
\raggedcolumns
\footnotesize
abraxane \\
alimta \\
ataxol \\
avastin \\
avstin \\
bevacizumab \\
caboplatin \\
cabotaxol \\
carbo \\
carboplat \\
carboplatin \\
carbotaxol \\
chemo \\
chemo therapy \\
chemo-rt \\
chemoembolization \\
chemorad \\
chemort \\
chemotherap \\
chemotherapeutic \\
chemotherapeutic \\
chemotherapies \\
chemotherapy \\
chemotherapy's \\
chemotheray \\
chmeo \\
cisplatin \\
cistoplatin \\
cytoreductive surgery \\
dcb \\
docetaxel \\
docetaxil \\
doxil \\
doxorubicin \\
elahere \\
etoposide \\
femara \\
gcb \\
gemcitabine \\
gemzar \\
hycamtin \\
ihcp \\
intraperitoneal hyperthemicchemoperfusion \\
intraperitonealhyperthemic chemoperfusion \\
koselugo \\
letrozole \\
liposomal doxorubicin \\
lynparza \\
mekinist \\
mirvetuximab soravtansine \\
navelbine \\
nintedanib \\
niraparib \\
nolvadex \\
olaparib \\
ovastat \\
paciltaxel \\
paclitaxel \\
paclitaxela \\
paraplatin \\
paxil \\
pazopanib \\
pemetrexed \\
platinol \\
platinum \\
pldc \\
rubraca \\
rucaparib \\
selumetinib \\
t/c \\
tamoxifen \\
tax \\
taxo \\
taxol \\
taxotere \\
tc \\
tc-bev \\
topotecan \\
trabectedin \\
trametinib \\
treosulfan \\
vargatef \\
vepesid \\
vinorelbine \\
votrient \\
yondelis \\
zejula \\

\normalsize

\section{Supplementary Figures and Tables}
\label{sec:supplementary_fig_tab}

\renewcommand{\thetable}{S\arabic{table}}
\setcounter{table}{0}

\begin{table}[ht]
\centering
\setlength{\tabcolsep}{4pt} 
\begin{tabular}{lrrr}
\toprule
\textbf{Cancer Type} & \textbf{\# Total} & \textbf{\# Annotated (\%)} \\
\midrule
Breast     & 14,234 & 216 (1.52\%) \\
Melanoma   & 9,279  & 523 (5.64\%) \\
Ovarian    & 16,782 & 523 (3.12\%) \\
\bottomrule
\end{tabular}
\caption{Proportion of sentences with annotations in the development set}
\label{tab:anno_stats}
\end{table}

\begin{table*}[ht]
\centering
\setlength{\tabcolsep}{6pt}
\begin{tabular}{l c ccc ccc}
\toprule
\multirow{2}{*}{\textbf{Cancer type}} & \multirow{2}{*}{\textbf{anno}} & \multicolumn{3}{c}{\textbf{Dictionary only}} & \multicolumn{3}{c}{\textbf{Dictionary + LLM verification}} \\
\cmidrule(lr){3-5}\cmidrule(lr){6-8}
& & \textbf{Precis.}& \textbf{Recall} & \textbf{F1} & \textbf{Precis.} & \textbf{Recall} & \textbf{F1} \\
\midrule
breast   & 0 & 0.9944 & 1.0000 & 0.9972 & 0.9967 & 1.0000 & 0.9984 \\
breast   & 1 & 0.7322 & 1.0000 & 0.8454 & 0.8244 & 1.0000 & 0.9038 \\
melanoma & 0 & 0.9853 & 1.0000 & 0.9926 & 0.9878 & 0.9998 & 0.9937 \\
melanoma & 1 & 0.8021 & 1.0000 & 0.8902 & 0.8296 & 0.9962 & 0.9053 \\
ovarian  & 0 & 0.9926 & 0.9998 & 0.9962 & 0.9927 & 0.9998 & 0.9963 \\
ovarian  & 1 & 0.8112 & 1.0000 & 0.8958 & 0.8150 & 0.9943 & 0.8958 \\
\bottomrule
\end{tabular}
\caption{Sentence-level tagging on the dev set: dictionary tagging vs.\ dictionary tagging with LLM verification. \texttt{anno\_indicator}=1 means the sentence has gold chemotherapy annotations.}
\label{tab:dict_llm_dev}
\end{table*}

\section{Example of Inconsistency Between Reasoning and Output}
\label{sec:example_of_inconsistency}
\noindent\textbf{Model thinking trace:}\\
<think> First, I need to check if "Tc" is a correct SACT. Wait, Tc-99m MDP is Technetium-99m Methane Diphosphate, which is a radiopharmaceutical used in bone scans. But SACT refers to systemic anticancer therapies like chemotherapy, endocrine therapy, etc. Radiopharmaceuticals are used for diagnostic purposes, not as part of cancer treatment. So "Tc" here is a false positive from the dictionary method. The user’s instruction says to remove incorrect tags caused by false positives. \textbf{Therefore, I should remove the <e>Tc</e> tag.} </think>\\

\noindent\textbf{Model output:}\\
TECHNIQUE: Whole body bone scan was obtained following intravenous administration of 26.5mCi of <e>Tc</e>-99m MDP.\\

\noindent\textbf{Expected corrected output:}\\
TECHNIQUE: Whole body bone scan was obtained following intravenous administration of 26.5mCi of Tc-99m MDP.\\

\noindent This example highlights a misalignment between the model’s reasoning and its generated output: although the reasoning correctly identifies ``Tc'' as a false positive (radiopharmaceutical rather than chemotherapy), the structured output still preserves the erroneous tag.

\end{document}